\documentclass[conference]{IEEEtran}
\IEEEoverridecommandlockouts
\usepackage{cite}
\usepackage{amsmath,amssymb,amsfonts}
\usepackage{algorithmic}
\usepackage{graphicx}
\usepackage{comment}
\usepackage{textcomp}
\usepackage{xcolor}
\usepackage{lipsum}
\usepackage{stfloats}
\usepackage{multicol}
\usepackage{graphicx}
\def\BibTeX{{\rm B\kern-.05em{\sc i\kern-.025em b}\kern-.08em
    T\kern-.1667em\lower.7ex\hbox{E}\kern-.125emX}}
\usepackage{multicol}
\usepackage[english]{babel}
\usepackage[autostyle]{csquotes}

\usepackage{geometry}
\begin{document}

\title{AI-Based Copyright Detection Of An Image In a Video Using Degree Of Similarity And Image Hashing}

\author{\IEEEauthorblockN{Ashutosh}
\IEEEauthorblockA{\text{Department of Electrical Engineering} \\
\text{Indian Institute of Technology} \\
Dharwad, India \\
190020008@iitdh.ac.in}
\and
\IEEEauthorblockN{Rahul Jashvantbhai Pandya}
\IEEEauthorblockA{\text{Department of Electrical Engineering} \\
\text{Indian Institute of Technology} \\
Dharwad, India \\
rpandya@iitdh.ac.in}
}

\IEEEpubid{\makebox[\columnwidth]{ \hfill} \hspace{\columnsep}\makebox[\columnwidth]{ }}
\maketitle
\IEEEpubidadjcol

\begin{abstract}
   The expanse of information available over the internet makes it difficult to identify whether a specific work is a replica or a duplication of a protected work, especially if we talk about visual representations. Strategies are planned to identify the utilization of the copyrighted image in a report. Still, we want to resolve the issue of involving a copyrighted image in a video and a calculation that could recognize the degree of similarity of the copyrighted picture utilized in the video, even for the pieces of the video that are not featured a lot and in the end perform characterization errands on those edges. Machine learning (ML) and artificial intelligence (AI) are vital to address this problem. Numerous associations have been creating different calculations to screen the identification of copyrighted work. This work means concentrating on those calculations, recognizing designs inside the information, and fabricating a more reasonable model for copyrighted image classification and detection. We have used different algorithms like- Image Processing, Convolutional Neural Networks (CNN), Image hashing, Etc.

\textit{Keywords}---Copyright; Artificial Intelligence(AI); Copyrighted Image; Convolutional Neural Network(CNN); Image processing; Degree of similarity; Image Hashing;

\end{abstract}

\section{Introduction}
Copyright is intended to preserve the original representation of a notion in a creative work, not the idea itself. A copyright is an authorized development that, generally for a particular period, empowers its owner to reproduce, send, change, show, and perform creative innovations. In certain jurisdictions, protected works should be fixed in an unmistakable structure. The innovative idea could appear as something melodic, creative, instructive, or abstract. It is frequently written by several authors, known as rights holders, who each have a different set of rights to use or license the work. Common examples of these rights include those related to reproduction, distribution, public performance, control over derivative works, and moral rights like attribution. Copyright can be used to secure a wide range of innovative, literary, or artistic ideas. Creative write-ups such as poems, theses, hymns, fictional characters' roles, theatre, and other literary creations may all be included in this category, as well as video content, choreography, songs, audio recordings, illustrations, paintings, artworks, and photographs, as well as software applications, broadcast media, and design patents, although specifics will vary by jurisdiction. In some jurisdictions, there may be overlapping or different rules that apply to graphic and industrial designs. Any type of property's owner has the right to determine how it is to be used, and anybody else can only use it legally with the owner's consent, which is sometimes granted through a license. However, the owner's use of the property must respect other people's legally recognized rights and interests. Therefore, the owner of a copyright-protected work has the right to determine how to use the work and to forbid unauthorized use of it\cite{Copyright}.

The literature currently available discusses protecting a work, mainly image/audio/video, from the inclusion of copyrighted work using watermarking techniques. An owner authenticity symbol (watermark) is inserted into the host signal using watermarking techniques, and the watermark data can subsequently be retrieved. The watermark data, which may or may not be visible, can include a set of binary data, a range of samples, or a single bit in the host signal. Other techniques involve image inpainting using neural architecture\cite{article1}. Image inpainting is the term for restoration techniques intended to naturally eliminate imperfection or undesired elements from an image so that a viewer who is not biased would mistake the outcome for the original image. Research has been done on the video copyright protection system through different mechanisms such as watermark pre-processing, watermark embedding, database management, Etc.

The current work tries to detect the use of copyrighted work, primarily photographs in a video, as the primary element of a frame in a video or is utilized in certain frames but is less highlighted. The following is done by measuring the degree of similarity\cite{article3} using the Oriented FAST and Rotated BRIEF(ORB)\cite{article} algorithm and Structural Similarity(SSIM). Based on the degree of similarity, we can set a threshold that, if crossed, will point to copyright detection. Perceptual image hashing also deals with the above problem and we use it to separate out similar images on comparing with the copyrighted image.

\section{System Model}
We address the various models put out in this paper, such as image classification, ORB feature matching, structural similarity, and image hashing, in this section. 

The proposed algorithm includes\\
1. Frames fragmentation from a video using OpenCV\cite{OpenCV} as test file\\
2. Level of similarity detection between copyrighted image and frame obtained.\\
3. Creation of a dataset out of frames obtained and applying
   image hashing to compare similar frames with copyrighted images and separate them out.

Initial input will come from a video that needs to be evaluated using a copyrighted image (test image). Then, this video is broken up into frames using two different methods: first, by pre-processing it and feeding it to the image hashing algorithm; second, by feeding video directly to the algorithm, frames are then compared side by side with the test image and a degree of similarity is displayed of each frame, which is used to detect copyrighted data.\\
Now, for finding the similarity between the features of the two images we will be using two ORB and SSIM. BRIEF descriptors are used by ORB, but they rotate poorly. Therefore, ORB rotates the BRIEF in accordance with the position of key points whereas the Structural Similarity Index (SSIM\cite{article2}) is a perceptual metric that measures the loss in image quality brought on by data transmission losses or other processing steps like data compression.

\subsection{ORB feature matching}
\enquote{ Oriented FAST and Rotated BRIEF, The well-known FAST keypoint detector and the BRIEF descriptor serve as the foundation for ORB. Both of these methods are appealing due to their high effectiveness and low expense.}\\
\enquote{FAST is an algorithm for identifying interest points in an image\cite{FAST}. BRIEF is a general-purpose feature point descriptor that can be combined with arbitrary detectors.}\cite{BRIEF}\\
The main contributions of ORB are, adding a quick and precise orientation component, Analysis of variance and correlation of oriented BRIEF features, Etc.

\begin{figure}[htbp]
    \includegraphics[scale=0.44]{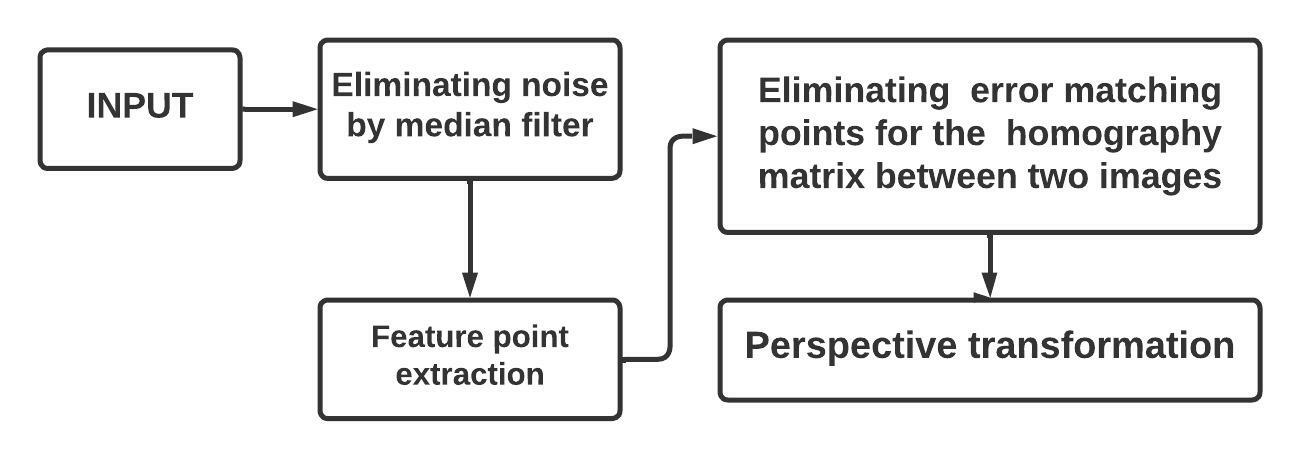}
    \caption{ORB feature matching algorithm}
    \end{figure}

\subsection{Structural Similarity(SSIM)}
\enquote{The SSIM is a perception-based model that incorporates crucial perceptual phenomena, such as luminance and contrast masking terms, and views image degradation as a perceived change in structural information. The distinction between these methods and others is that they estimate absolute errors, unlike MSE or PSNR.} \cite{ASSIM}.\\

\begin{figure}[htbp]
    \includegraphics[scale=0.4]{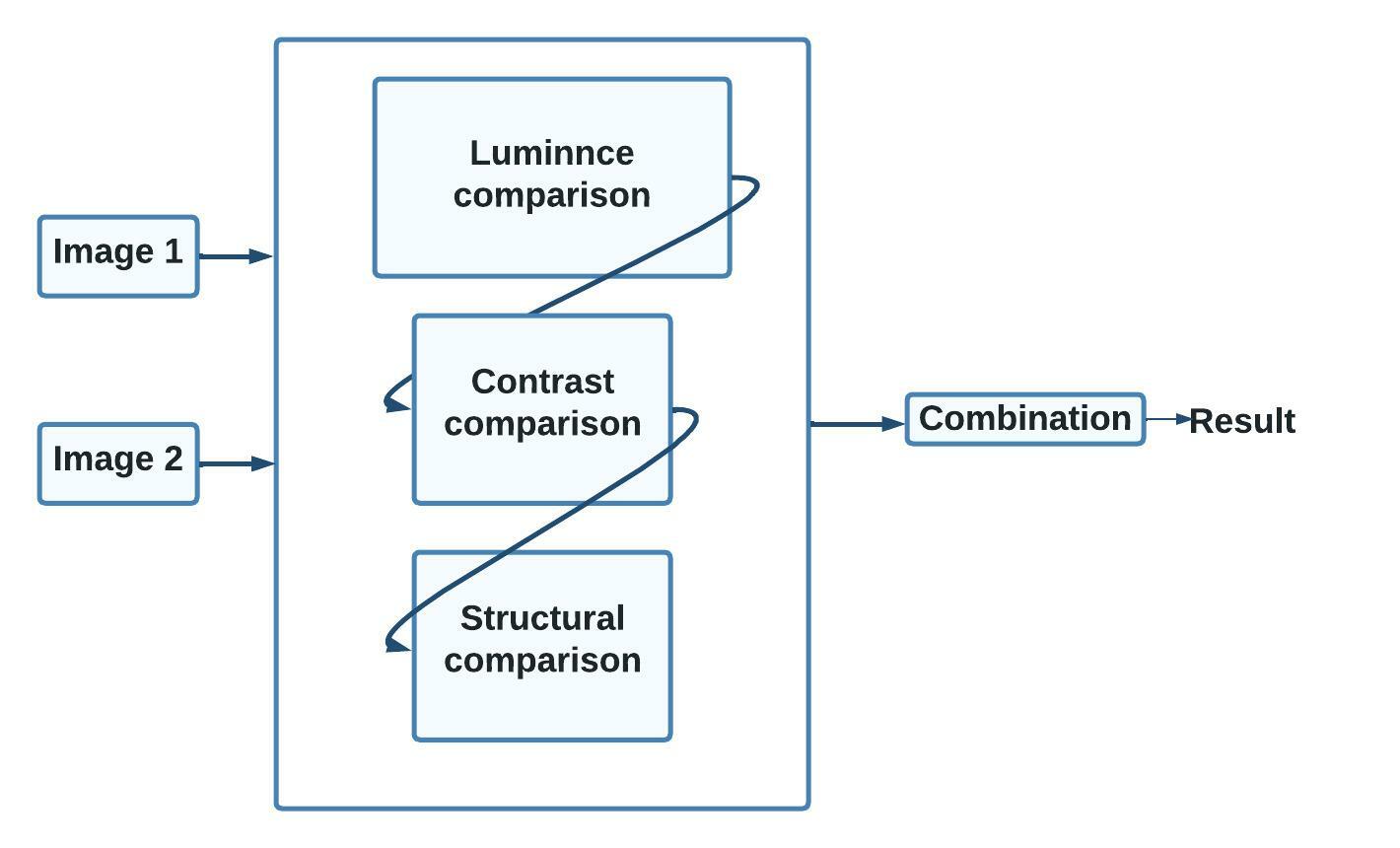}
    \caption{Structural Similarity (x and y inputs to be compared)\cite{}}
    \end{figure}

We are using the above two techniques to measure the degree of similarity.

\subsection{Image Hashing}
\enquote{A group of algorithms called perceptual image hashing create content-based image hashes.}\\
Using an algorithm to give each image a distinct hash value is known as Image Hashing.

\begin{figure}[htbp]
    \includegraphics[scale=0.8]{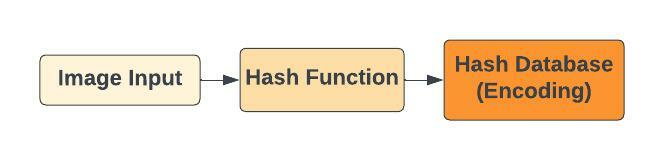}
    \caption{Image hashing}
    \end{figure}

\section{Methodology}
\subsection{Computation of degree of similarity}
The proposed Classification model is only intended for research and has no intention of generating a final product. This model compares features of two, a test image (copyrighted) and frames fetched from a video, and suggests a value for the level of similarity; when this value exceeds a threshold, image copyright is indicated. A video is used as an input to check for a copyrighted image's use. The flow of the model is well shown in \enquote{Figure 4}.\\
To properly judge intensity, the obtained frames are pre-processed in the appropriate color format. We used a test video shot in a garden. Now, Figure 10 is a test image which is a butterfly, we used to detect its appearance in the video. According to the simulation results, we discovered that the highest similarity measure, by the ORB algorithm is 0.097 and, by the SSIM algorithm is 0.301 which is for the frame where the butterfly is coming in the clip, even though this value is too low suggesting significant less amount of feature matching, so we can infer that the given image is not used in the video and there won't be any copyright detection.\\
As a first step in developing our approach, we wrote separate functions to determine the ORB similarity and the SSIM similarity measures. Then, we created a loop where each frame is captured, the similarity measures are calculated for each frame by executing the aforementioned methods, and the attributes of each frame are compared to those of the test image.

\begin{figure}[h]
    \centering
    \includegraphics[scale=0.33]{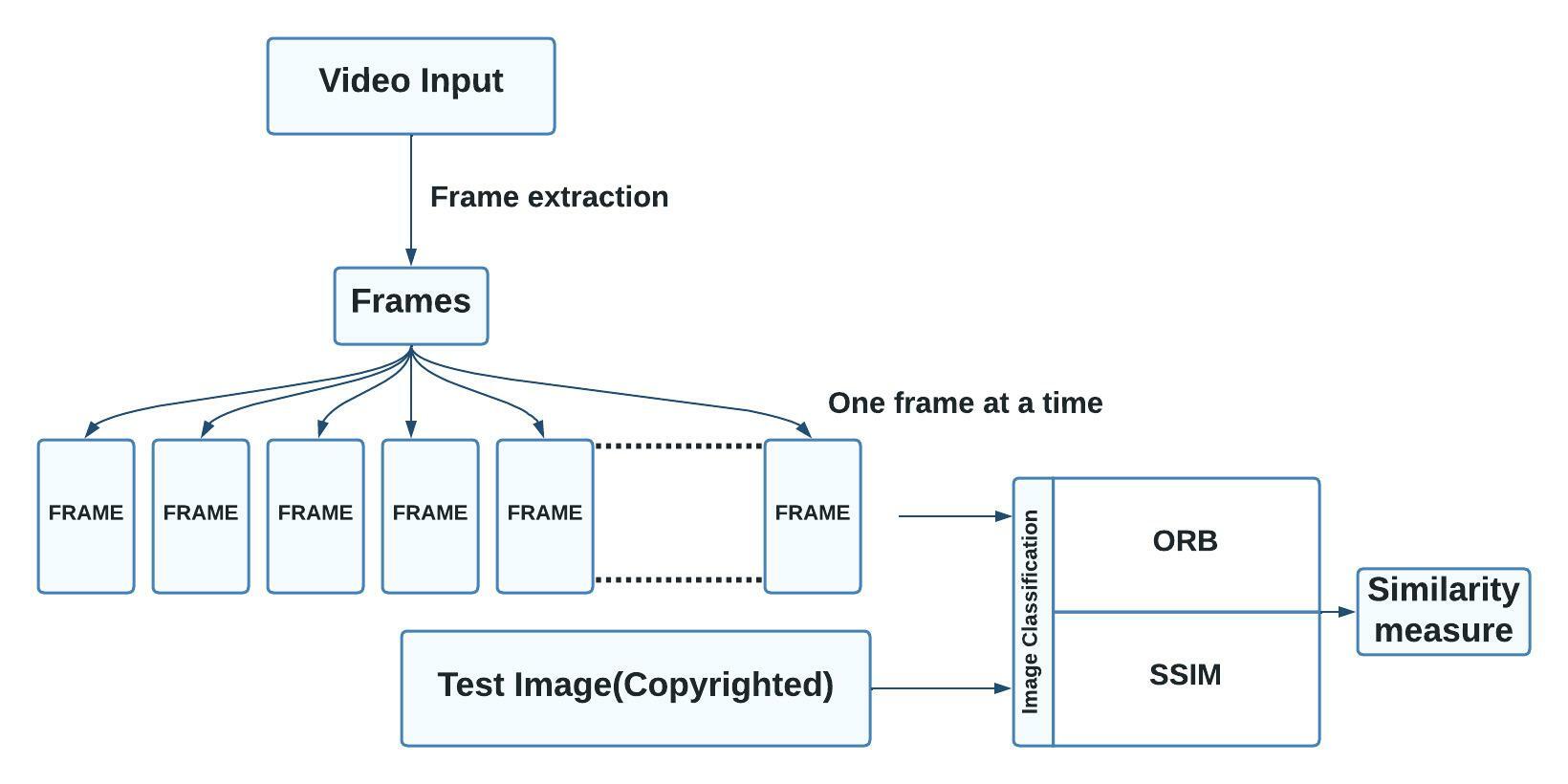}
    \caption{Working mechanism}
    \end{figure}
\subsubsection{ORB function}
ORB uses three main steps: feature point extraction, feature point descriptor generation, and feature point mapping.
The enhanced FAST (features from accelerated segment test) technique is used by the ORB algorithm to find feature points. According to the theory, a pixel is more likely to be a corner point if it differs greatly from its surrounding pixels. Say,\\ 
N = pixel on which we are analyzing.\\
$I_{N}$ = Intensity of color for Nth pixel.\\
K = Range deciding value.\\
Using pixel N as the center, choose 16 other pixels from the circle, and compare the grey values of the 16 selected pixels to those of pixel N. A pixel N can be regarded as a feature point if the brightness of a series of subsequent "n" points on the chosen circle is more than $I_{N}$ + K or less than $I_{N}$ - K\cite{article}.
The ORB algorithm enhances the baseline FAST algorithm, which determines the Harris response values for the baseline FAST corner points. Harris response value calculation,\cite{HR}
\\
Harris response,\[ R = det(P)-k(trace(P))^{2} ,\\ k:=[0.04,0.06]\]\\
P is a 2X2 matrix windowed over the summation of intensity measurements.

\begin{figure}[htbp]
    \centering
    \includegraphics[scale=0.36]{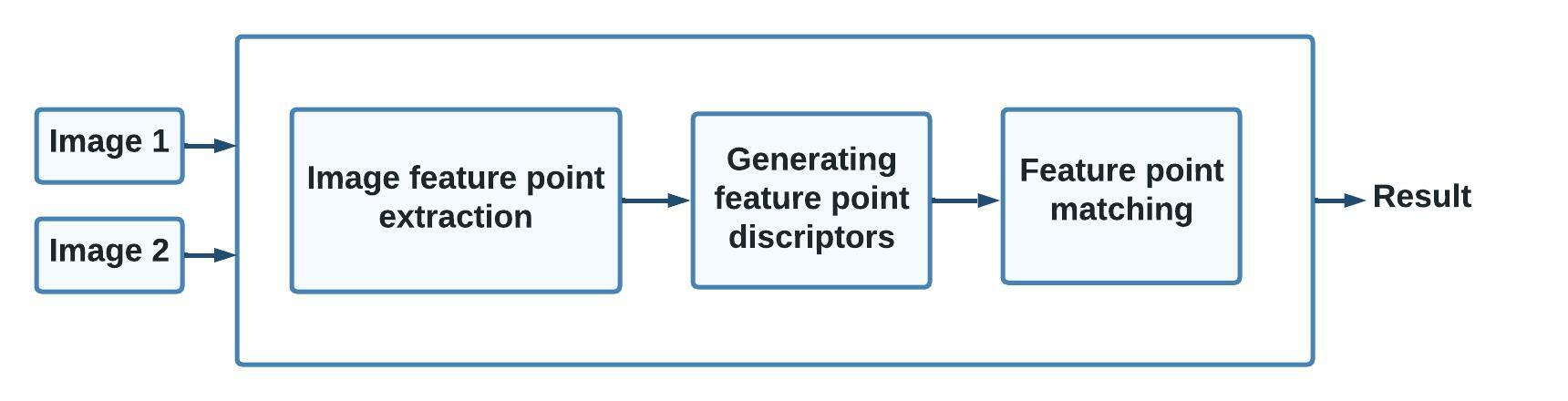}
    \caption{ORB functionality}
    \end{figure}

\subsubsection{SSIM function\cite{SS}}
Reference, Figure(2), \\
L = Brightness comparison function.\\
\[L = (2*u_{x}*u_{y}+q)/(u_{x}^{2}+u_{y}^{2}+q)\]\\
u = Luminance (averaging over all the pixel values).\\
\[u = (1/N)\sum x_{i}\]
q = $(K*l)^{2}$, l is the dynamic range for pixel values, and K is a normal constant.\\
C = contrast comparison function.\\
\[C = (2*s_{x}*s_{y}+q)/(s_{x}^{2}+s_{y}^{2}+q)\]\\
s = Contrast (Standard deviation of pixel values).\\
\[s = ((1/N-1)\sum (x_{i}-u_{x})^{2})^{1/2}\]
q = $(K*l)^{2}$, The dynamic range of pixel values is defined by l, and K is a normal constant.
S = Structure comparison function.\\
\[S = (s_{xy}+q)/(s_{x}*s_{y}+q)\]\\
\[s_{xy} = ((1/N-1)\sum (x_{i}-u_{x})*(y_{i}-u_{y}))\]
q = $(K*l)^{2}$, l is the dynamic range for pixel values, and K is a normal constant.
\[SSIM score = L^{a}*C^{b}*S^{c}, a,b,c > 0 \]a,b,c denote the relative importance of each.

\begin{figure}[htbp]
    \centering
    \includegraphics[scale=0.4]{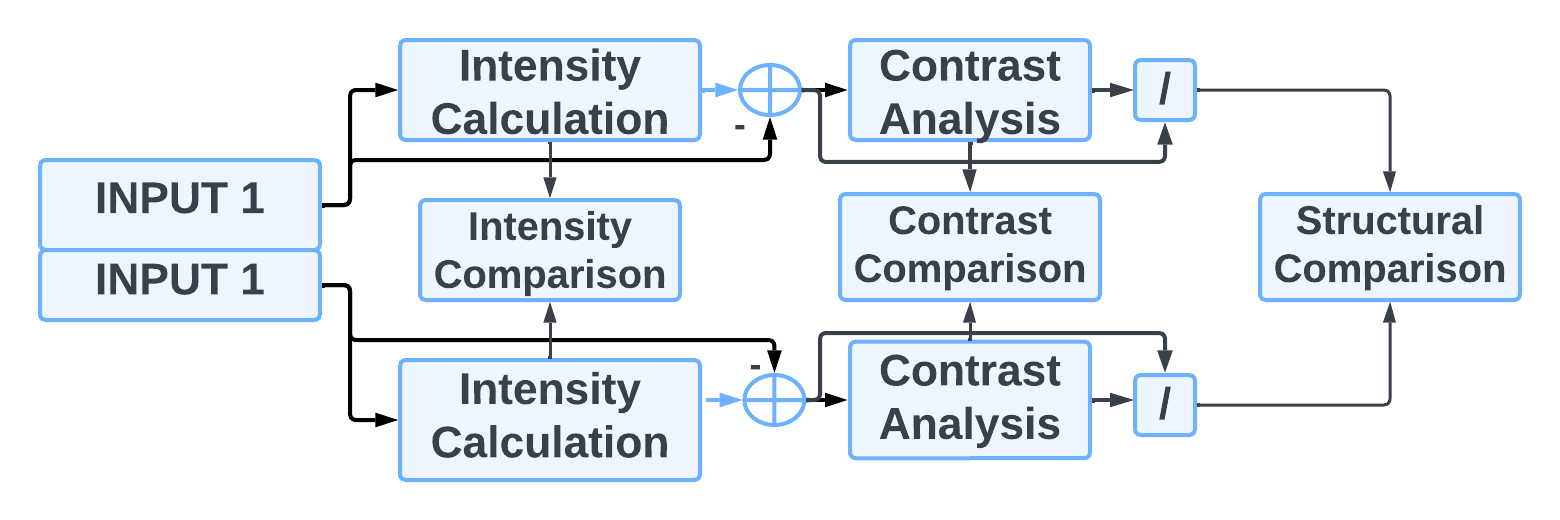}
    \caption{Similarity Measure}
    \end{figure}
\subsection{Image Hashing}
Image hashing is the most widely recognized estimation strategy to consign a specific hash worth to a picture. Copy duplicates of the image all have a similar hash esteem. Thus, it is sometimes called a 'digital unique finger impression' for a photograph.
\begin{figure}[htbp]
    \centering
    \includegraphics[scale=0.18]{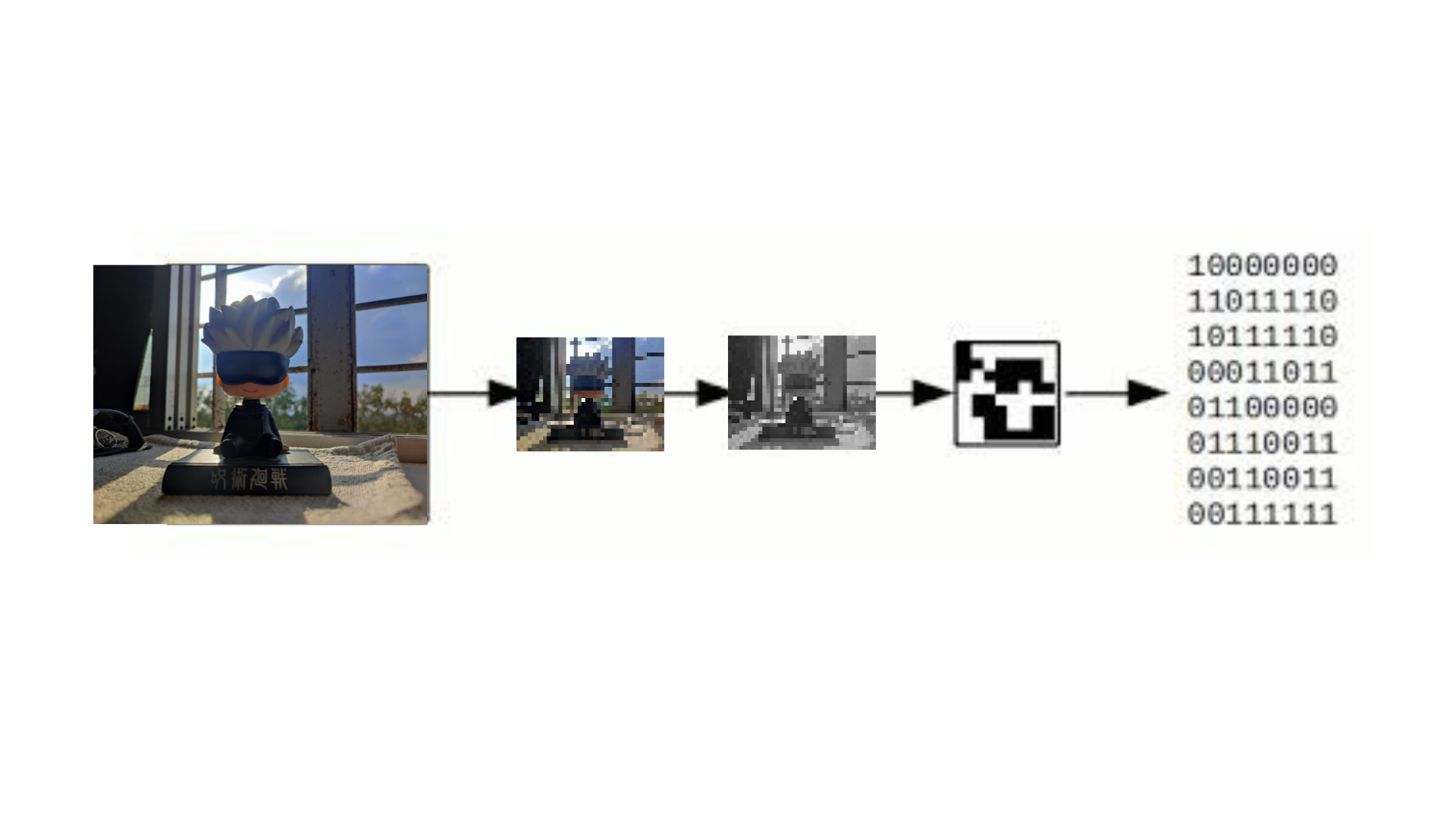}
    \caption{Image Hash generation}
    \end{figure}
\begin{figure}[htbp]
    \centering
    \includegraphics[scale=0.35]{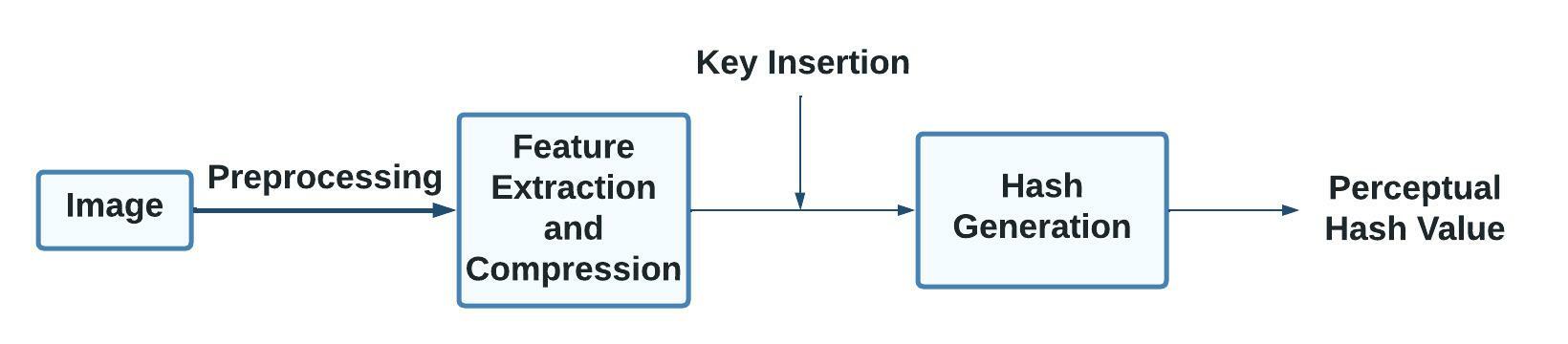}
    \caption{Hashing System}
    \end{figure}

\section{SIMULATION RESULT}
For Copyright detection, we have used a video with a total frame count of 560 but picked up frames evenly at an interval of 10 frames. The test image is that of a car. In the figure below, we have a snippet shown of some of the frames:

\begin{figure}[h]
    \centering
    \includegraphics[scale=0.5]{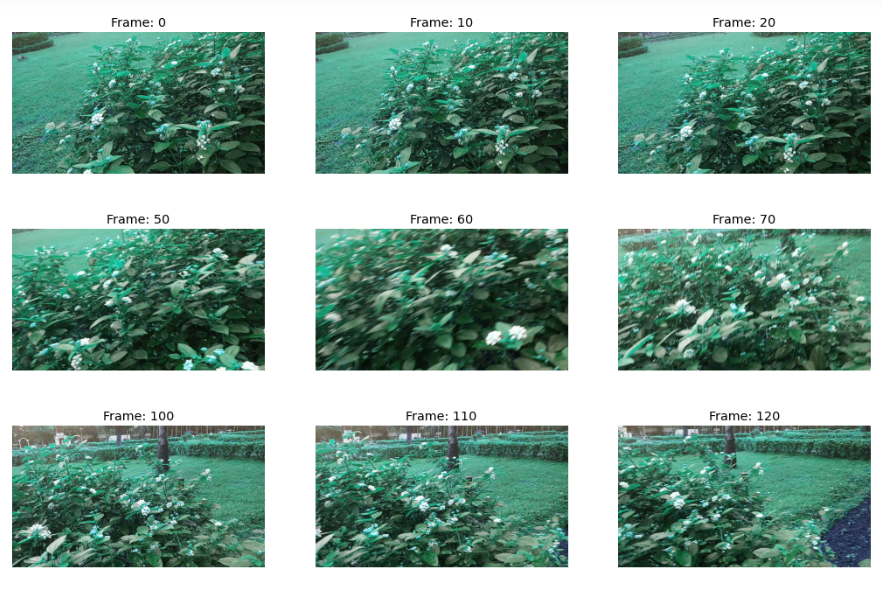}
    \caption{Frames}
    \end{figure}

Also, please find the test image \enquote{figure 10} we have used.

\begin{figure}[htbp]
    \centering
    \includegraphics[scale=0.8]{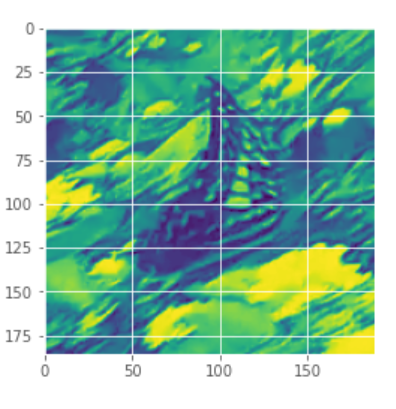}
    \caption{Test Image}
    \end{figure}

ORB similarity measure between each "frame" and "test image" as a result of our algorithm is as follows,\\
\
[\ 0.045, 0.024, 0.018, 0.016, 0.009, 0.029, 0.061, 0.097
, 0.028, 0.007, 0.022, 0.025, 0.035, 0.0, 0.016, 0.033, 0.033, 0.0, 0.079, 0.058, 0.047, 0.015, 0.008, 0.007, 0.007, 0.008]\ \\
Note -  here 1.0 means identical image, and the graphical representation is attached in figure - 11.\\
\\

\begin{figure}[h]
    \centering
    \includegraphics[scale=0.8]{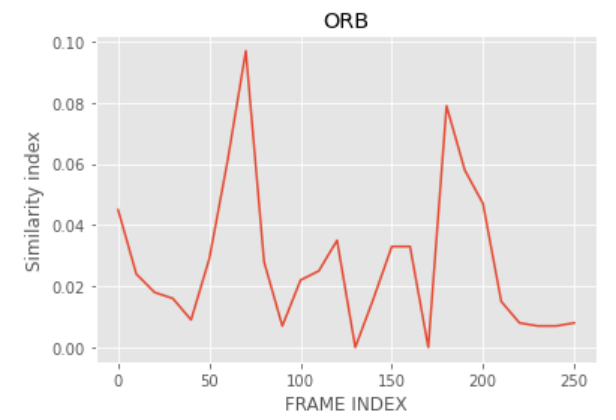}
    \caption{Curve represents ORB similarity measure}
    \end{figure}

SSIM similarity measure between each "frame" and "test image" as a result of our algorithm is as follows,\
[\ 0.106, 0.165, 0.176, 0.228, 0.171, 0.238, 0.301, 0.250, 0.159, 0.121, 0.12, 0.137, 0.157, 0.0927, 0.15, 0.172, 0.141, 0.134, 0.264, 0.238, 0.183, 0.143, 0.106, 0.0928, 0.095]\ \\
Note -  here 1.0 means identical image, and the graphical representation is attached in figure - 12.\\

\begin{figure}[h]
    \centering
    \includegraphics[scale=0.7]{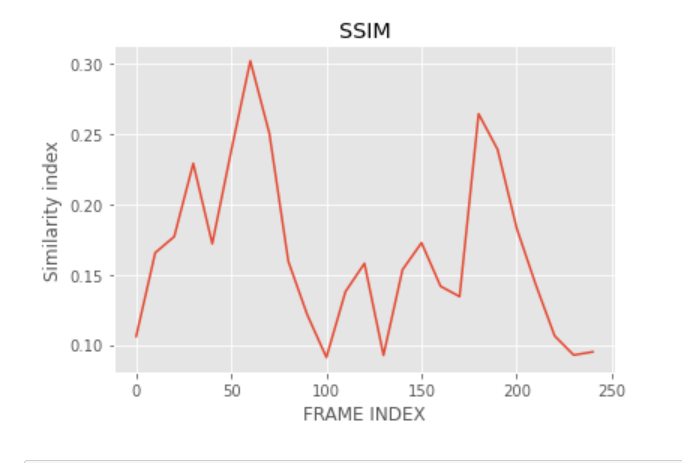}
    \caption{Curve represents SSIM similarity measure}
    \end{figure}

We use a set of codes for image hashing to form a dataset by dividing the video into the required number of frames. Then our image hashing algorithm segregates the unique images and detects the similar-looking images. This is done by comparing the hash values of two images. If it crosses a threshold(refers to a boundary value that would decide whether an image is copied or not), it pushes a similar image into a bin folder and separates the unique images as the working is shown in figure-13.

\begin{figure}[h]
    \centering
    \includegraphics[scale=0.7]{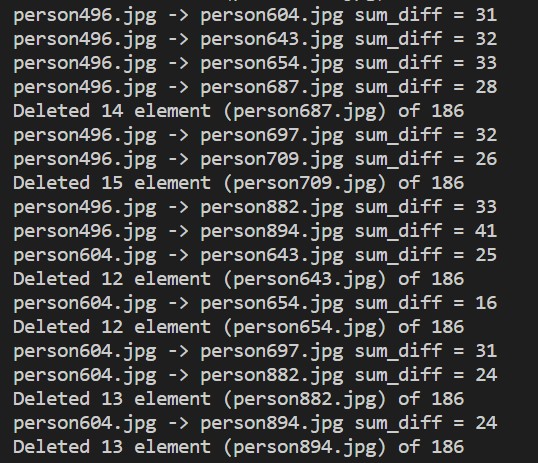}
    \caption{Comparison results and images transfer based on similarity }
    \end{figure}

\section{Conclusion}
Based on the user's level of threshold selection, the copyright detection model can determine the similarity measure of different frames in a video with a given test image. As a result, they are addressing the issue of image copyright detection in videos. \\
The explained indexes act as a base to the creation for the high level dataset which acts as the labelled training data for the multidimensional labelled training for machine learning algorithms, Hence automating the process.\\
As per figure - 11 and figure - 12, at the end it depends on user to user as of what is the extent up to which there is an allowance for the copyrighted data usage, if the barriers are high then the test image in above run is not used in the video, if similarity measures are too low, then we may infer no copyright issues are there presently and vice-versa.

\bibliographystyle{IEEEtran}
\bibliography{references.bib}
\end{document}